\documentclass[runningheads]{llncs}

\usepackage[mobile]{eccv}

\usepackage{eccvabbrv}

\usepackage{graphicx}
\usepackage{booktabs}
\usepackage{marvosym}
\usepackage[accsupp]{axessibility}  

\usepackage[pagebackref,breaklinks,colorlinks,citecolor=eccvblue]{hyperref}

\usepackage{orcidlink}
\usepackage{multirow}
\usepackage{makecell}
\usepackage{pythonhighlight}

\begin{document}

\title{DriveDreamer-2: LLM-Enhanced World Models for Diverse Driving Video Generation} 

\titlerunning{DriveDreamer-2}

\author{Guosheng Zhao \inst{1*}\and
Xiaofeng Wang \inst{1*} \and
Zheng Zhu\inst{2*}\textsuperscript{\Letter}\and
Xinze Chen\inst{2}\and\\
Guan Huang\inst{2}\and
Xiaoyi Bao\inst{1}\and
Xingang Wang\inst{1}\textsuperscript{\Letter} 
}
\authorrunning{G. Zhao, X. Wang et al.}

\institute{$^\text{1 }$Institute of Automation, Chinese Academy of Sciences~~$^\text{2 }$GigaAI\\
\small{Project Page: \url{https://drivedreamer2.github.io}}}


\maketitle
\vspace{-1em}
\begin{figure}[ht]
\centering
\subfloat[User-customized driving video generation.]{
\includegraphics[width=0.95\textwidth]{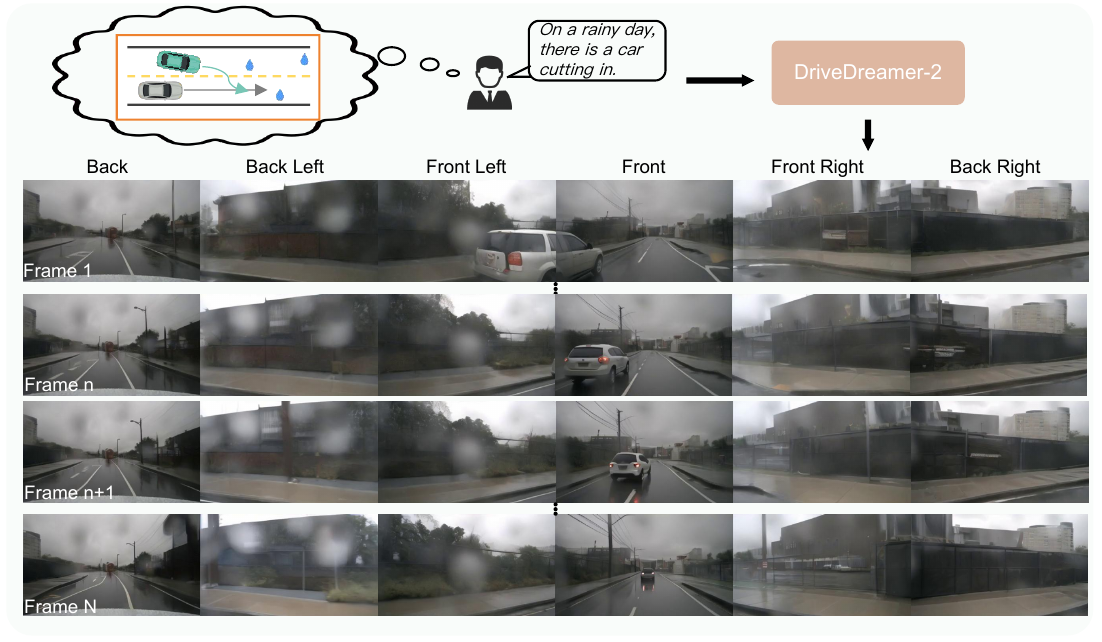}
\label{fig1a}}
\newline
\subfloat[Generated video quality comparison and improvement in the downstream task.]{
\includegraphics[width=0.95\textwidth]{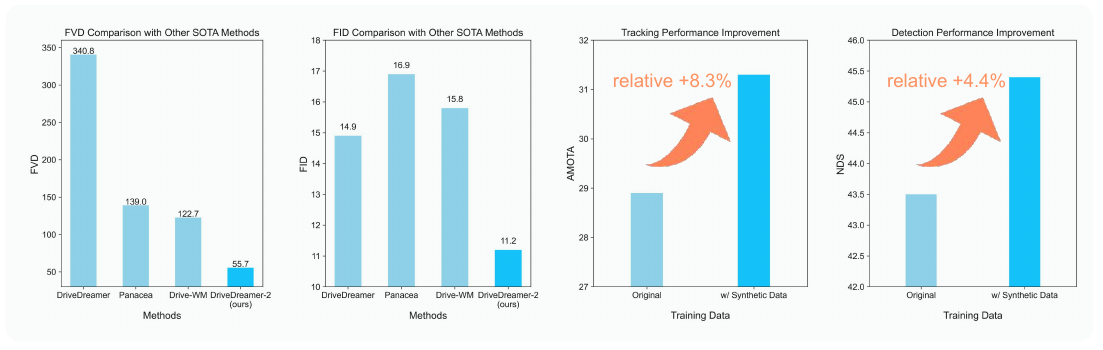} \label{fig1b}
}

\caption{\textit{DriveDreamer-2} demonstrates powerful capabilities in generating multi-view driving videos. \textit{DriveDreamer-2} can produce driving videos based on user descriptions, which improves the diversity of the synthetic data. Besides, the generation quality of \textit{DriveDreamer-2} surpasses other state-of-the-art methods and effectively enhances downstream tasks. }
\label{fig1}
\end{figure}


\vspace{-3.5em}
\begin{abstract}
World models have demonstrated superiority in autonomous driving, particularly in the generation of multi-view driving videos. However, significant challenges still exist in generating customized driving videos. In this paper, we propose \textit{DriveDreamer-2}, which builds upon the framework of DriveDreamer and incorporates a Large Language Model (LLM) to generate user-defined driving videos. Specifically, an LLM interface is initially incorporated to convert a user's query into agent trajectories. Subsequently, a HDMap, adhering to traffic regulations, is generated based on the trajectories. Ultimately, we propose the  Unified Multi-View Model to enhance temporal and spatial coherence in the generated driving videos. \textit{DriveDreamer-2} is the first world model to generate customized driving videos, it can generate uncommon driving videos (e.g., vehicles abruptly cut in) in a user-friendly manner. Besides, experimental results demonstrate that the generated videos enhance the training of driving perception methods (e.g., 3D detection and tracking). Furthermore, video generation quality of \textit{DriveDreamer-2} surpasses other state-of-the-art methods, showcasing FID and FVD scores of 11.2 and 55.7, representing relative improvements of $\sim$30\% and $\sim$50\%.

  
\end{abstract}

\section{Introduction}
\label{sec:intro}

World models for autonomous driving \cite{hu2023gaia,jia2023adriver,wang2023drivedreamer,wang2023driving} have drawn extensive attention from both the industry and academia in recent years. 
Benefiting from their excellent predictive capabilities, autonomous driving world models facilitate the generation of diverse driving videos, encompassing even long-tail scenarios. The generated driving videos can be utilized to enhance the training of various driving perception approaches, proving highly beneficial for practical applications in autonomous driving.



World modeling in autonomous driving presents a formidable challenge due to its inherent complexity and large sampling space. Early approaches \cite{mile,sem2} mitigate these problems by incorporating world modeling within the Bird’s Eye View (BEV) semantic segmentation space. However, these methods primarily explore world models in simulated autonomous driving environments. In the recent evolution of autonomous driving technologies, there has been a substantial leap forward in the development of world models. This progress has been propelled by the utilization of cutting-edge diffusion models  \cite{df1,df2,df3,glide,df5,df6}, exemplified by notable contributions such as DriveDreamer \cite{wang2023drivedreamer}, Drive-WM \cite{wang2023driving}, MagicDrive \cite{magicdrive}, Panacea \cite{wen2023panacea}, and the integration of large language models like GAIA-1 \cite{hu2023gaia}, ADriver-I \cite{jia2023adriver}. These sophisticated models have played a pivotal role in pushing the boundaries of world modeling capabilities, enabling researchers and engineers to delve into increasingly intricate and realistic driving scenarios.
However, it is important to note that a majority of these methods rely heavily on structured information (e.g., 3D boxes, HDMaps, and optical flow) or real-world image frames as conditions. This dependence not only constrains interactivity but also limits the diversity of generated videos.

To tackle the aforementioned challenges, we propose \textit{DriveDreamer-2}, which is the first world model to generate diverse driving videos in a user-friendly manner. In contrast to previous methods \cite{wang2023drivedreamer,wang2023driving,magicdrive} that rely on structured conditions either from specific datasets or sophisticated annotations, \textit{DriveDreamer-2} emphasizes generating customized driving videos by simulating various traffic conditions with user-friendly text prompts. 
Specifically, the traffic simulation task has been disentangled into the generation of foreground conditions (trajectories of the ego-car and other agents) and background conditions (HDMaps of lane boundary, lane divider, and pedestrian crossing). For foreground generation, a functional library is constructed to finetune a Large Language Model (LLM), enabling it to generate agent trajectories based on user text input.
For background conditions, we propose the HDMap generator that employs a diffusion model to simulate road structures. In this process, the previously generated agent trajectories are involved as conditional inputs, which allows the HDMap generator to learn the associations between foreground and background conditions in driving scenes.
Building upon the generated traffic structured conditions, we employ the DriveDreamer \cite{wang2023drivedreamer} framework to generate multi-view driving videos. It is noted that we introduce the Unified Multi-view Video Model (UniMVM) within the DriveDreamer framework, which is designed to unify both intra-view and cross-view spatial consistency, enhancing the overall temporal and spatial coherence in the generated driving videos.

Extensive experiment results show that \textit{DriveDreamer-2} is capable of producing diverse user-customized videos, including uncommon scenarios where vehicles abruptly cut in (depicted in Fig.~~\ref{fig1}). Besides, \textit{DriveDreamer-2} can generate high-quality driving videos with an FID of 11.2 and FVD of 55.7, relatively improving previous best-performing methods by $\sim$30\% and $\sim$50\%. Furthermore, experiments are conducted to verify that driving videos generated by \textit{DriveDreamer-2} can enhance the training of various autonomous driving perception methods, where the performance of detection and tracking are relatively improved by $\sim$4\% and $\sim$8\%.

The main contributions of this paper can be summarized as follows: 

\begin{itemize}
    \item We present \textit{DriveDreamer-2}, which is the first world model to generate diverse driving videos in a user-friendly manner.
    \item We propose a traffic simulation pipeline employing only text prompts as input, which can be utilized to generate diverse traffic conditions for driving video generation.
    \item UniMVM is presented to seamlessly integrate intra-view and cross-view spatial consistency, elevating the overall temporal and spatial coherence within the generated driving videos.
    \item Extensive experiments are conducted to show that \textit{DriveDreamer-2} can craft diverse customized driving videos. Besides, \textit{DriveDreamer-2} enhances the FID and FVD by $\sim$30\% and $\sim$50\% compared to previous best-performing methods. Moreover, the driving videos generated by \textit{DriveDreamer-2} enhance the training of various driving perception methods.
    
\end{itemize}

\vspace{-2em}
\section{Related Works}
\subsection{World Models}

The primary objective of world methods is to establish dynamic environmental models, endowing agents with predictive capabilities for the future. In the early exploration, Variational Autoencoders (VAE) \cite{vae} and Long Short-Term Memory (LSTM) \cite{lstm} are employed to capture transition dynamics and rendering functionality, showcasing remarkable success across diverse applications \cite{dreamv1,dreamv2,wm2,gamegan,worldmodel,wm1,wm2,wm3,wm4,wm5}. Constructing driving world models poses distinctive challenges, primarily arising from the high sample complexity inherent in real-world driving tasks \cite{e2esurvey}. To address these challenges, ISO-Dream \cite{isodream} introduces an explicit disentanglement of visual dynamics into controllable and uncontrollable states. MILE \cite{mile} strategically incorporates world modeling within the Bird's Eye View (BEV) semantic segmentation space. Recently, DriveDreamer \cite{wang2023drivedreamer}, GAIA-1 \cite{hu2023gaia}, ADriver-I \cite{jia2023adriver}, and Drive-WM \cite{wang2023driving} have explored the training of driving world models in the real world, leveraging powerful diffusion models or natural language models. However, most of these methods heavily depend on structured information (e.g., 3D boxes, HDMaps, and optical flow) as conditions. This dependency not only constrains interactivity but also limits generation diversity.
\subsection{Video Generation}

Video generation and prediction are pivotal techniques for understanding the visual world. In the early stages of video generation, methods like Variational Autoencoders (VAEs) \cite{videopred2, vae2}, flow-based model \cite{flow1}, and Generative Adversarial Networks (GANs) \cite{gan1,gan2,gan3,gan4} are explored. Language models \cite{ar1,ar2,ar3,ar4,wang2024worlddreamer,hong2022cogvideo,villegas2022phenaki,kondratyuk2023videopoet} are also employed for intricate visual dynamics modeling.
Recent advancements have seen diffusion models \cite{df1,df2,df3,glide,df5,df6} extending their influence to video generation. Notably, video diffusion models \cite{vdm1,vdm2,vdm3,vdm4,vdm5,vdm6,blattmann2023stable} exhibit superior capabilities in generating high-quality videos with realistic frames and smooth transitions, offering enhanced controllability. These models adapt seamlessly to various input conditions, including text, canny, sketch, semantic maps, and depth maps. In the realm of autonomous driving, \textit{DriveDreamer-2} leverages powerful diffusion models for learning visual dynamics.


\subsection{Traffic Simulation}
Driving simulators stand as a cornerstone in self-driving development, aiming to offer a controlled environment to mimic real-world conditions. LCTGen\cite{tan2023language} utilizes an LLM to encode detailed language descriptions to a vector and subsequently employs a generator to produce corresponding simulated scenarios. This method requires highly detailed language descriptions, including information such as the speed and orientation of agents. TrafficGen\cite{feng2023trafficgen} comprehends the inherent relationships within traffic scenarios, enabling the generation of diverse and legitimate traffic flows within the same map. CTG\cite{zhong2023guided} generates traffic simulations by employing manually designed loss functions that adhere to traffic constraints. CTG++\cite{zhong2023language} further extends CTG by utilizing GPT-4\cite{openai2023gpt} to convert user language descriptions into a loss function, which guides the scene-level conditional diffusion model to generate the corresponding scenario. In \textit{DriveDreamer-2}, we construct a functional library to finetune the LLM to achieve a user-friendly text-to-traffic simulation, which eliminates intricate loss design or complex text prompt inputs.

\begin{figure}[t]
\centering
\includegraphics[width=\textwidth]{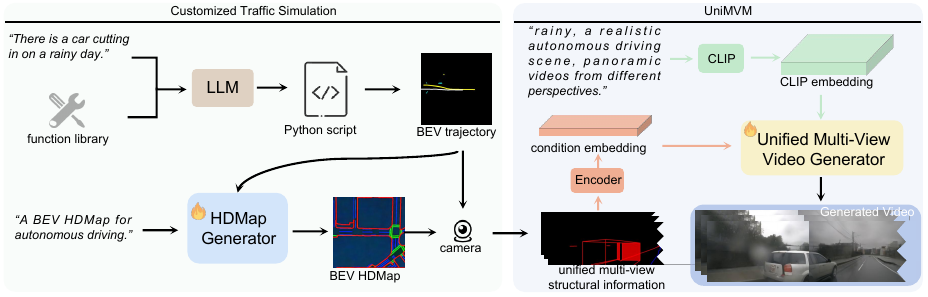}
\caption{The overall framework of \textit{DriveDreamer-2} involves initially generating agent trajectories according to the user query, followed by producing a realistic HDMap, and finally generating multi-view driving videos.}
\label{fig_framework}
\end{figure}
\section{DriveDreamer-2}
Fig.~~\ref{fig_framework} illustrates the overall framework of \textit{DriveDreamer-2}. A customized traffic simulation is first proposed to generate foreground agent trajectories and background HDMaps. Specifically, \textit{DriveDreamer-2} utilizes a finetuned LLM to translate user prompts into agent trajectories, and the HDMap generator is then introduced to simulate road structures using the generated trajectories as conditions. Leveraging the customized traffic simulation pipeline, \textit{DriveDreamer-2} is capable of generating diverse structured conditions for the subsequent video generation. Building upon the architecture of DriveDreamer \cite{wang2023drivedreamer}, the UniMVM framework is proposed to unify both intra-view and cross-view spatial consistency, thereby enhancing the overall temporal and spatial coherence in the generated driving videos. In the subsequent sections, we delve into the details of the customized traffic simulation and the UniMVM framework.

\subsection{Customized Traffic Simulation}
In the proposed customized traffic simulation pipeline, a trajectory-generation function library is constructed to finetune the LLM, which facilitates transferring user prompts into diverse agent trajectories, encompassing maneuvers such as cut-ins and U-turns. Additionally, the pipeline incorporates the HDMap generator to simulate the background road structures. During this phase, the previously generated agent trajectories serve as conditional inputs, ensuring that the resulting HDMap adheres to traffic constraints. In the following, we elaborate on the finetuning process of the LLM and the framework of the HDMap generator.

\begin{figure}[t]
\centering
\includegraphics[width=\textwidth]{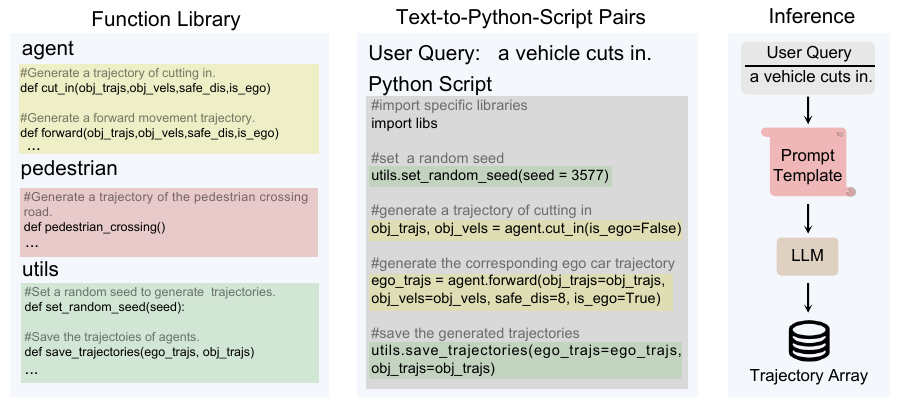}
\caption{The overview of customized trajectory generation. Initially, we leverage the established function library to assemble Text-to-Python-Script pairs. Subsequently, the constructed dataset is employed to finetune LLM. Finally, the customized trajectories is generated by LLM based on user query.}
\vspace{-1em}
\label{fig_script}
\end{figure}

\noindent \textbf{Finetuning LLM for Trajectory Generation} Previous traffic simulation methods \cite{mao2023gpt,zhong2023language,zhong2023guided} necessitate the intricate specification of parameters, involving details such as the agent's speed, position, acceleration, and mission goal. To simplify this intricate process, we propose to finetune LLM with the constructed trajectory-generation function library, allowing for the efficient transformation of user-friendly language inputs into comprehensive traffic simulation scenarios.
As depicted in Fig.~\ref{fig_script}, the constructed function library encompasses 18 functions, including agent functions (steering, constant speed, acceleration, and braking), pedestrian functions (walking direction and speed), and other utility functions such as saving trajectories. Building upon these functions, Text-to-Python-Scripts pairs are manually curated for finetuning LLM (GPT-3.5). The scripts include a range of fundamental scenarios such as lane-changing, overtaking, following other vehicles, and executing U-turns. Additionally, we encompass more uncommon scenarios like pedestrians abruptly crossing, and vehicles cutting into the lane. Taking the user input \textit{a vehicle cuts in} as an example, the corresponding script involves the following steps: initially generating a trajectory of cutting in (\texttt{agent.cut\_in()}), followed by generating the corresponding ego car trajectory (\texttt{agent.forward()}), and ultimately utilizing the saving function from utilities to directly output the trajectory of the ego-car and other agents in array format. For additional details, please refer to the supplementary materials.  In the inference phase, we follow \cite{mao2023gpt} to expand prompt inputs to a pre-defined template, and the finetuned LLM can directly output the trajectory array.




\begin{figure}[t]
\centering
\includegraphics[width=\textwidth]{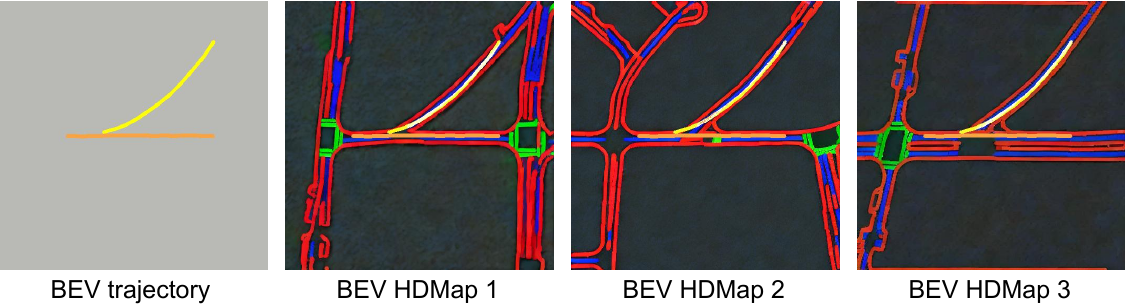}
\caption{The proposed HDMap generator can generate diverse BEV HDMaps based on the same BEV trajectory input. The \textcolor{orange}{orange} and \textcolor{yellow}{yellow} colors represent the motion trajectories of the ego car and other vehicles, respectively. The \textcolor{red}{red} color indicates road boundaries, the \textcolor{blue}{blue} color represents lane dividers, and the \textcolor{green}{green} color signifies pedestrian crossings.}
\label{fig_hdmap}
\end{figure}

\noindent \textbf{HDMap Generation} 
A comprehensive traffic simulation not only entails the trajectories of foreground agents but also necessitates the generation of background HDMap elements such as lanes and pedestrian crosswalks. Therefore, the HDMap generator is proposed to ensure the background elements do not conflict with the foreground trajectories. In the HDMap generator, we formulate the background elements generation as a conditional image generation problem, where the conditional input is the BEV trajectory map ${\cal T}_b \in {\cal R}^{3\times H_b\times W_b}$, and the target is the BEV HDMap ${\cal H}_b \in {\cal R}^{3\times H_b\times W_b}$. Different from previous conditional image generation approaches \cite{controlnet,gligen} that predominantly rely on outline conditions (edges, depths, boxes, segmentation maps), the proposed HDMap generator explores the correlations between the foreground and background traffic elements. Specifically, the HDMap generator is constructed upon an image-generation diffusion model. To train the generator, we curate a trajectory-to-HDMap dataset ${\cal D}=\{{\cal T}_b,{\cal H}_b\}$. In the trajectory map, distinct colors are assigned to represent different agent categories. Meanwhile, the target HDMap comprises three channels, representing lane boundaries, lane dividers, and pedestrian crossings, respectively. Within the HDMap generator, we employ stacks of 2D convolution layers to incorporate the trajectory map condition. The resulting feature maps $C_{\cal T}$ are then seamlessly integrated into the diffusion model using \cite{controlnet} (see supplement for additional architectural details). In the training stage, the diffusion forward process gradually adds noise $\epsilon$ to the latent feature ${\cal Z}_0$, resulting in the noisy latent feature ${\cal Z}_{T_b}$. Then we train $\epsilon_\theta$ to predict the noise we added, and the HDMap generator $\phi$ is optimized via


\begin{equation}
   \min_{\phi} {\cal L} = {\mathbb E}_{{\cal Z}_0,\epsilon \sim {\cal N}({\textbf{0,I}}),t,c} \left[\Vert\epsilon-\epsilon_\theta({\cal Z}_t,t,c)\Vert_2^2\right],
\end{equation}
where time step t is uniformly sampled from $[1,T_b]$. As shown in Fig.~~\ref{fig_hdmap}, leveraging the proposed HDMap generator allows us to generate diverse HDMaps based on the same trajectory conditions. It is noteworthy that the generated HDMaps not only adhere to traffic constraints (lane boundaries positioned on either side of lane dividers, and pedestrian crossings at intersections) but also seamlessly integrate with trajectories.


\subsection{UniMVM}

Utilizing structured information generated by the customized traffic simulation, multi-view driving videos can be generated via the framework of DriveDreamer \cite{wang2023drivedreamer}. However, the view-wise attention introduced in previous methods \cite{wang2023drivedreamer,bevcontrol} can not guarantee multi-view consistency. To mitigate this problem, \cite{wang2023driving,wen2023panacea,drivediff} employ image or video conditions to generate multi-view driving videos. While this approach enhances consistency between different views, it comes at the expense of reduced generation efficiency and diversity. In \textit{DriveDreamer-2}, we introduce the UniMVM within the DriveDreamer framework. The UniMVM is designed to unify the generation of multi-view driving videos both with and without adjacent view conditions, which ensures temporal and spatial coherence without compromising generation speed and diversity.

\begin{figure}[t]
\centering
\includegraphics[width=\textwidth]{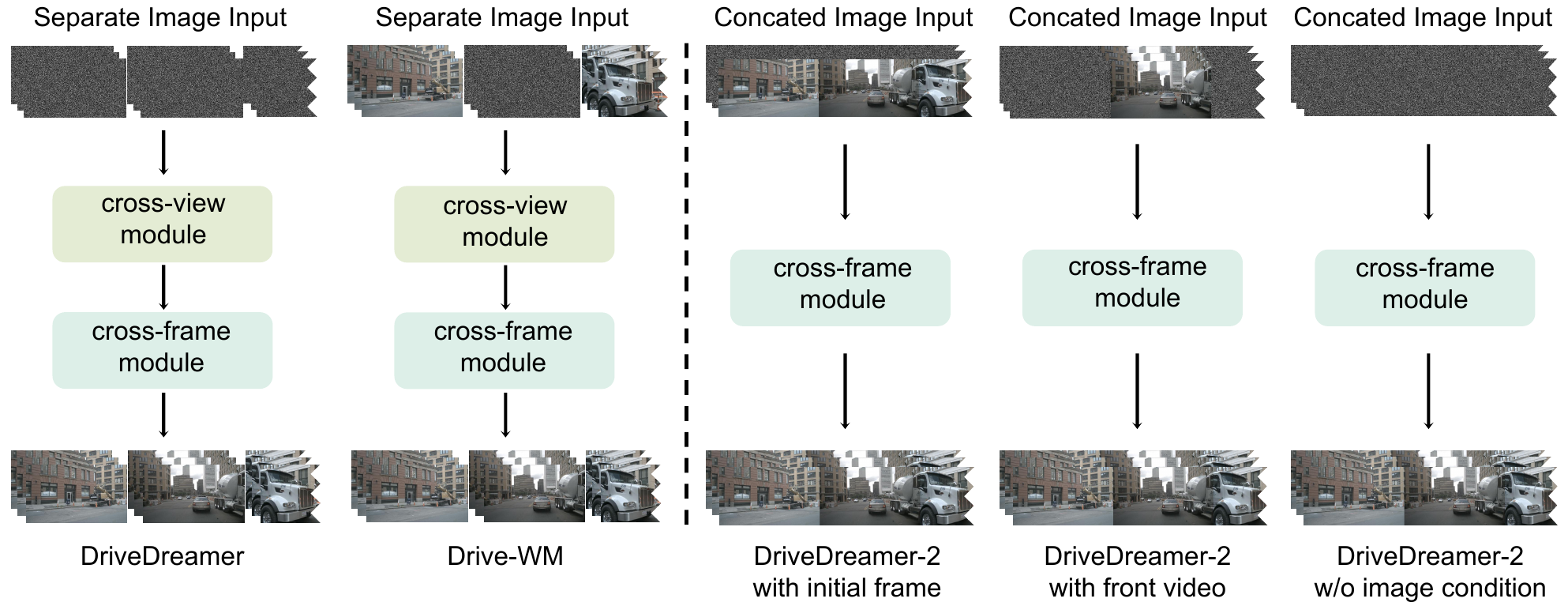}
\caption{The comparison of multi-view video generation paradigms. All structural conditions and text prompts are omitted here to emphasize the distinctions between our UniMVM and previous methods. By adjusting the mask, UniMVM can generate videos conditioned on the initial frame, front view video, and without image input.}
\vspace{-1em}
\label{fig_unimvm}
\end{figure}


\noindent \textbf{Formulation} In multi-view video dataset $p_{data}$, ${\rm x}\in {\cal R}^{K\times T \times3\times H \times W}$ is a sequence of $T$ images with $K$ views, with height $H$ and width $W$. Let ${\rm x}_i$ denote the sample of $i$-th view, then the multi-view video joint distribution $p({\rm x_1, ...,  \rm x_K})$ can be obtained by \cite{wang2023driving}:
\begin{equation} \label{eq2}
p({\rm x_1, ...,  \rm x_K}) =
 p({\rm x}_1)p({\rm x}_{2}|{\rm x}_1)...
 p\left(\left. {\rm x}_{K} \right| {\rm x}_{1},{\rm x}_2...,{\rm x}_{K-1}\right).
\end{equation}
Eq.~\ref{eq2} indicates that adjacent view videos can be expanded with multiple generation steps, which is inefficient. In the proposed UniMVM, we draw inspiration from the Eq.~\ref{eq2} to expand the view. However, unlike Drive-WM \cite{wang2023driving} which requires the independent generation of views, UniMVM unifies multiple views as a complete patch. Specifically, we concatenate the multi-view video in the order of \{FL, F, FR, BR, B, BL\}\footnote{F:Front, L: Left, R: Right, B: Back.} to obtain the spatially unified image  ${\rm x}'\in {\cal R}^{T\times3\times H\times KW}$. Then we can obtain the multi-view driving video distribution $p({\rm x}')$:

\begin{equation}
  p({\rm x}')=p({\rm x}'\cdot (1-m),{\rm x}'\cdot m)=p({\rm x}'\cdot m)p({\rm x}'\cdot (1-m)|{\rm x}'\cdot m),
\end{equation}
where $m$ represents the mask of one of the all views. 
As shown in Fig.~\ref{fig_unimvm}, we compare the paradigm of UniMVM with that of DriveDreamer \cite{wang2023drivedreamer} and Drive-WM \cite{wang2023driving}. In contrast to these counterparts, UniMVM unifies multiple views into a complete patch for video generation without introducing cross-view parameters. Furthermore, various driving video generation tasks can be accomplished via adjusting the mask $m$. Specifically, when $m$ is set to mask future $T-1$ frames, UniMVM enables future video prediction based on the input of the first frame. Configuring $m$ to mask \{FL, FR, BR, B, BL\} views empowers UniMVM to achieve multi-view video outpainting, leveraging a front-view video input. Furthermore,  UniMVM can generate multi-view videos when $m$ is set to mask all video frames, and both quantitative and qualitative experiments verify that UniMVM is capable of generating temporally and spatially coherent videos with enhanced efficiency and diversity.




\noindent \textbf{Video Generation} Based on the UniMVM formulation, driving videos can be generated within the framework of DriveDreamer \cite{wang2023drivedreamer}. Specifically, our approach first unify the traffic structured conditions, which results in sequences of HDMaps $\{{\cal H}_i\}_{i=0}^{N-1}\in {\cal R}^{N\times 3\times H\times KW}$ and 3D boxes $\{{\cal B}_i\}_{i=0}^{N-1}\in {\cal R}^{N\times C\times H\times KW}$ ($N$ is the frame number of video clip, and $C$ is the category number). Note that sequences of 3D boxes can be derived from agent trajectories, and the sizes of 3D boxes are determined based on the respective agent category.
Unlike DriveDreamer, the 3D box conditions in \textit{DriveDreamer-2} no longer rely on position embedding and category embedding. Instead, the boxes are directly projected onto the image plane, functioning as a control condition. This approach eliminates introducing additional control parameters as in \cite{wang2023drivedreamer}.
We adopt three encoders to embed HDMaps, 3D boxes, and image frames into latent space features $y_{\cal H}$, $y_{\cal B}$, and $y_{\cal I}$. Then we concatenate the spatially aligned conditions $y_{\cal H}$, $y_{\cal B}$ with ${\cal Z}_t$ to obtain the feature input ${\cal Z}_{in}$, where ${\cal Z}_t$ is the noisy latent feature generated from $y_{\cal I}$ by the forward diffusion process. 
For the training of the video generator, all parameters are optimized via denoising score matching \cite{edm} (see supplement for details).

\vspace{-1em}
\section{Experiment}
\subsection{Experiment Details}

\noindent\textbf{Dataset.}
The training dataset is derived from the nuScenes dataset \cite{nusc}, consisting of 700 training videos and 150 validation videos. Each video encompasses approximately 20 seconds of recorded footage, captured by six surround-view cameras. With a frame rate of 12Hz, this accumulates to around 1 million video frames available for training.
Following \cite{wang2023drivedreamer,wang2023we}, we preprocess the nuScenes dataset to calculate 12Hz annotations. Specifically, HDMaps and 3D boxes are transformed to BEV perspective to train HDMap generation, and these annotations are projected to pixel coordinate to train video generation.

\noindent \textbf{Training.} For agent trajectory generation, we employ GPT-3.5 as the LLM. Subsequently, we utilize the constructed text-to-script dataset to finetune GPT-3.5 into an LLM with specialized trajectory generation knowledge. The proposed HDMap generator is built upon SD2.1 \cite{df6} with the ControlNet parameters \cite{controlnet} being trainable. The HDMap generator undergoes 55K training iterations with a batch size of 24 and a resolution of $512\times 512$. For the video generator, we harness the powerful video generation capabilities of SVD\cite{blattmann2023stable}, and all the parameters are finetuned. During the training of the video generator, the mode is trained for 200K iterations with a batch size of 1, a video frame length of $N=8$, a view number of $K=6$, and a spatial size of $256\times 448$. All the experiments are conducted on NVIDIA A800 (80GB) GPUs, and we use the AdamW optimizer\cite{kingma2014adam} with a learning rate $5\times10^{-5}$.


\noindent \textbf{Evaluation.} Extensive qualitative and quantitative experiments are conducted to assess \textit{DriveDreamer-2}. For qualitative experiments,  we visualize customized driving video generation to validate that \textit{DriveDreamer-2} can produce diverse driving videos in a user-friendly manner. Additionally, visualization comparisons are conducted between UniMVM and other generative paradigms to demonstrate that \textit{DriveDreamer-2} excels in generating temporally and spatially coherent videos.
 For quantitative experiments,  the frame-wise Fréchet Inception Distance (FID) \cite{parmar2022aliased} and Fréchet Video Distance (FVD) \cite{unterthiner2018towards} are utilized as metrics. Besides, StreamPETR\cite{wang2023exploring}, building upon a ResNet-50 \cite{he2016deep} backbone, is trained at the same resolution of $256\times 448$ to evaluate the improvements of 3D object detection and multi-object tracking achieved by our generated results. More details can be found in the supplementary material.

\subsection{User-Customized Driving Video Generation}

\begin{figure*}[t]
\centering
\includegraphics[width=\textwidth]{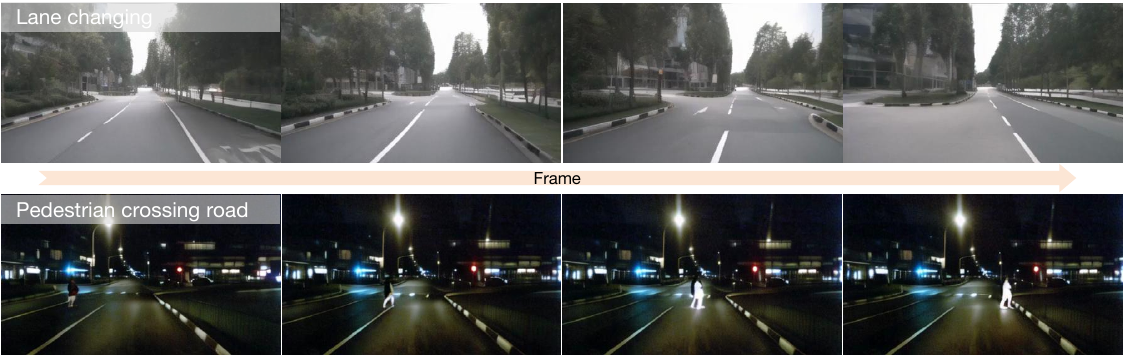}
\caption{User-customized driving videos generated by \textit{DriveDreamer-2}. The top row depicts a scene where the ego car changes lanes, while the bottom row shows an unexpected pedestrian crossing the road at night. }
\label{fig_cus}
\end{figure*}

\textit{DriveDreamer-2} offers a user-friendly interface for generating driving videos. As depicted in Fig.~\ref{fig1a}, users are only required to input a text prompt (e.g., \textit{on a rainy day, there is a car cut in}). Then \textit{DriveDreamer-2} produces multi-view driving videos aligned with the text input. Fig.~\ref{fig_cus} illustrates another two customized driving videos. The upper one depicts the process of the ego car changing lanes to the left during the daytime. The lower one showcases an unexpected pedestrian crossing the road at night, prompting the ego car to brake to avoid the collision. Notably, the generated videos demonstrate an exceptional level of realism, where we can even observe the reflection of high beams on the pedestrian. 

\begin{table}[h]
    \begin{center}
    \caption{Comparison of the generation quality on nuScenes validation set. \dag~denotes that the corresponding conditions are generated.}\label{tab_fvd}
    \setlength{\tabcolsep}{4mm}{
    \begin{tabular}{cccc}
        \hline 
        Method&Conditions& FID$\downarrow$& FVD$\downarrow$ \\
        \hline 
        DriveDreamer \cite{wang2023drivedreamer}&-&26.8&353.2\\
        \textit{DriveDreamer-2}&-&\textbf{25.0}&\textbf{105.1}\\
        \hline
        Drive-WM \cite{wang2023driving}&3-view videos\dag&\textbf{15.8}&122.7\\
        \textit{DriveDreamer-2}&1-view video&18.4&\textbf{74.9}\\
        \hline
        DriveDreamer \cite{wang2023drivedreamer}&1st-frame multi-view image&14.9&340.8\\
        Drivingdiffusion \cite{drivediff}&1st-frame multi-view image\dag&15.8&332.0\\
        Panacea \cite{wen2023panacea}&1st-frame multi-view image\dag&16.9&139.0\\
        \textit{DriveDreamer-2}&1st-frame multi-view image&\textbf{11.2}&\textbf{55.7}\\
        \hline
    \end{tabular}}
    \vspace{-3em}
    \end{center}
\end{table}

\subsection{Quality Evaluation of Generated Videos}


To verify the video generation quality, we compare \textit{DriveDreamer-2} with various driving video generation approaches on the nuScenes validation set.
For a fair comparison, we conducted evaluations under three different experimental settings—without image condition, with video condition, and with first frame multi-view image condition. Additionally, Drive-WM \cite{wang2023driving}, DrivingDiffusion \cite{drivediff} and Panacea \cite{wen2023panacea} adopt a two-stage pipeline, first generating visual conditions and then generating videos. 
The experimental results, as shown in Tab.~\ref{tab_fvd}, indicate that \textit{DriveDreamer-2} consistently achieves high-quality evaluation outcomes across all three settings. Specifically, in the absence of the image condition, \textit{DriveDreamer-2} attains an FID of 25.0 and an FVD of 105.1, showcasing a significant improvement over DriveDreamer \cite{wang2023drivedreamer}. Moreover, despite being limited to a single-view video condition, \textit{DriveDreamer-2} exhibits a 39\% relative improvement in FVD compared to DriveWM \cite{wang2023driving}, which utilizes a three-view video condition. Furthermore, when provided with the first-frame multi-view image condition, \textit{DriveDreamer-2} achieves an FID of \textbf{11.2} and an FVD of \textbf{55.7}, surpassing all previous methods by a considerable margin.
\begin{table}[ht]
    \begin{center}
    \caption{Comparison involving data augmentation using synthetic data on 3D object detection.}\label{tab_detction}
    \setlength{\tabcolsep}{1.7mm}{
    \begin{tabular}{ccc ccc ccc }
        \hline  
        Image size&Initial frame&Real&Generated&mAP$\uparrow$ &mAOE$\downarrow$&mAVE$\downarrow$ &NDS$\uparrow$ \\
        \hline
        \multirow{3}*{256$\times$448}&-&\checkmark&-&31.7&67.9&33.0&43.5\\
        ~&\checkmark&\checkmark&\checkmark&32.6&61.7&\textbf{29.7}&45.2 \\
        ~&-&\checkmark&\checkmark&\textbf{32.9}&\textbf{61.5}&30.4&\textbf{45.4} \\
        \hline
    \end{tabular}}
    \vspace{-1em}
    \end{center}
\end{table}

\begin{table}[t]
    \begin{center}
    \caption{Comparison involving data augmentation using synthetic data on multi-object tracking.}\label{tab_mot}
    \setlength{\tabcolsep}{1.7mm}{
    \begin{tabular}{c ccc ccc }
        \hline  
        Image size&Initial frame&Real&Generated&AMOTA$\uparrow$ &AMOTP$\downarrow$&IDS$\downarrow$ \\
        
        \hline
        \multirow{3}*{256$\times$448}&-&\checkmark&-&28.9&1.419&687\\
        ~&\checkmark&\checkmark&\checkmark&31.2&1.396&\textbf{542}\\
        ~&-&\checkmark&\checkmark&\textbf{31.3}&\textbf{1.387}&593\\
        \hline
    \end{tabular}}
    \vspace{-3em}
    \end{center}
\end{table}

To further validate the quality of the generated data, we employ the generated driving videos to enhance the training of 3D object detection and multi-object tracking. Specifically, we employ all structured conditions in the nuScenes training set to generate driving videos, which are combined with real videos to train StreamPETR \cite{wang2023exploring} on downstream tasks. The experiment results are in Tab.~\ref{tab_detction} and Tab.~\ref{tab_mot}. When the initial frame is used as a condition, the generated videos prove to be effective in enhancing the performance of downstream tasks. The 3D detection metrics, mAP and NDS, show a relative improvement of 2.8\% and 3.9\%, respectively. Besides, the tracking metrics, AMOTA and AMOTP, exhibit a relative enhancement of 8.0\% and 1.6\%. 
Furthermore, \textit{DriveDreamer-2} is capable of generating high-quality driving videos even in the absence of image conditions. Removing the image condition leads to increased diversity in the generated content, consequently further improving performance metrics for downstream tasks. Specifically, the mAP and NDS for 3D detection show a relative improvement of 3.8\% and 4.4\%, respectively, while the AMOTA and AMOTP for tracking tasks exhibit enhancements of 8.3\% and 2.3\%, compared to the baseline.
\vspace{-1em}
\begin{figure}[ht]
\centering
\includegraphics[width=\textwidth]{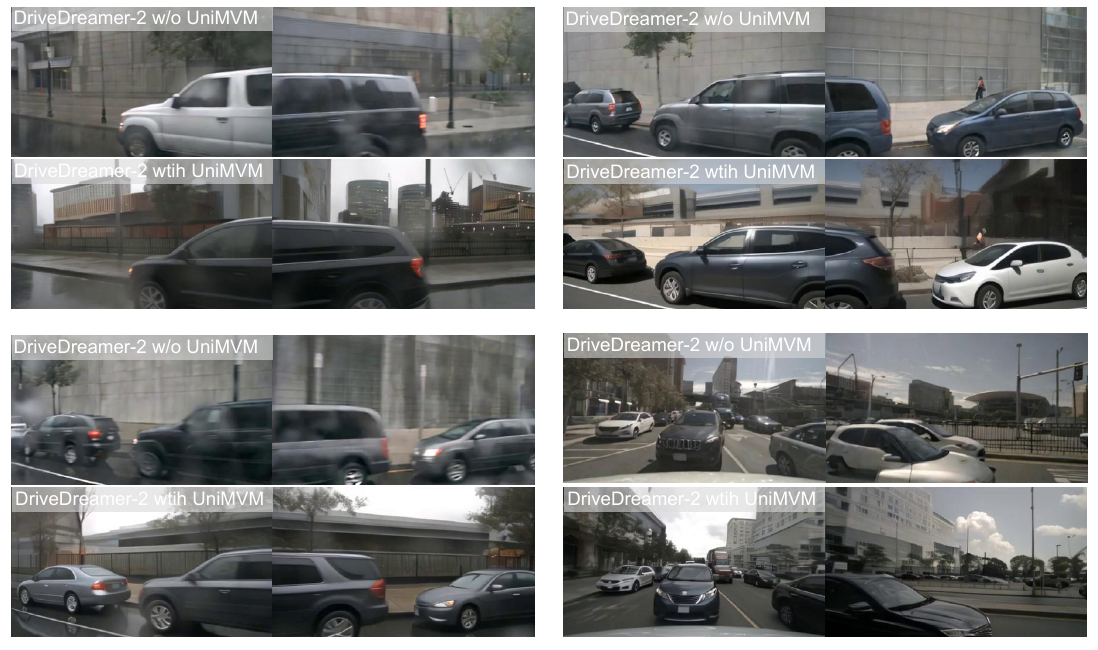}
\caption{Visualization comparison between \textit{DriveDreamer-2} generation with and without UniMVM. The upper part depicts generation without UniMVM, while the lower part illustrates generation with UniMVM. It is evident that the inclusion of UniMVM results in higher multi-view consistency in the generated content.}
    \vspace{-1em}
\label{fig_mv_1}
\end{figure}

\begin{figure}[ht]
\centering
\includegraphics[width=\textwidth]{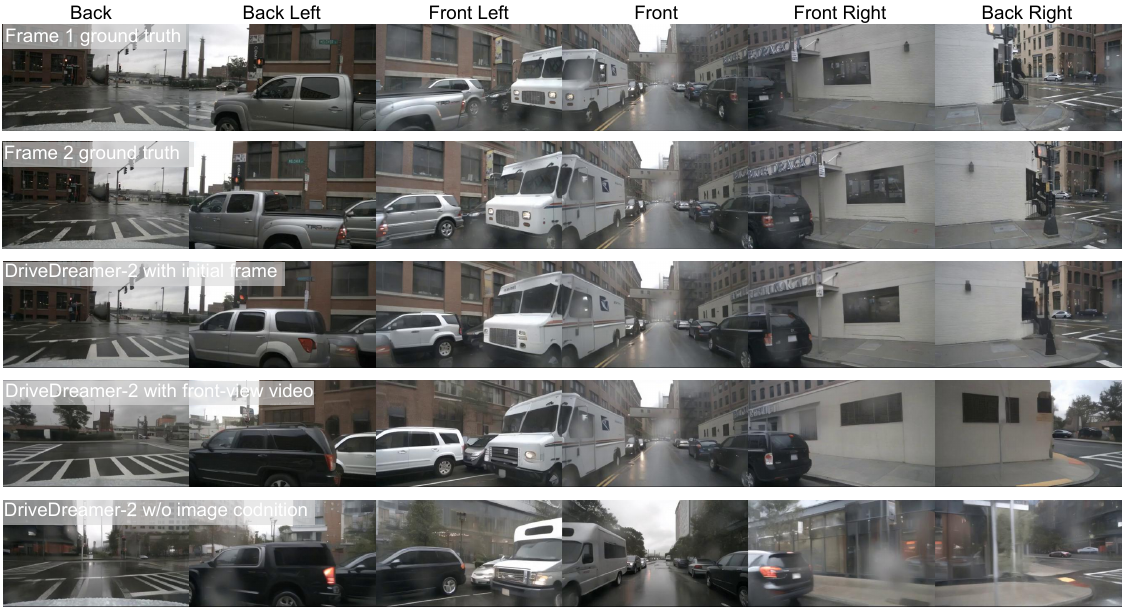}
\caption{Visualization comparison with different conditions. Under different image conditions, \textit{DriveDreamer-2} produces videos with a high level of multi-view consistency.  Using the 1st frame image (row 1) as a condition, the diversity of the generated second frame (row 3) is constrained, which closely resembles the ground truth second frame (row 2). Employing the front-view video as a condition results in increased generation diversity (row 4), only the front-view image aligns closely with the ground truth (row 2). Strikingly, in the absence of image as a condition, \textit{DriveDreamer-2} produces the highest diversity (row 5). The generated colors of the cars and the street backgrounds differ significantly from the ground truth (row 2).}
    \vspace{-2em}
\label{fig_mv}
\end{figure}

\subsection{Ablation Study} \label{4.4} 
We conduct an ablation study to investigate the effect of diffusion backbone and the proposed UniMVM, and the results are in Tab.~\ref{tab_ablation}.
Compared to SD1.4\cite{df6} used in DriveDreamer, SVD\cite{blattmann2023stable} provides richer prior knowledge of videos, resulting in 17.2 FID and 94.6 FVD. The introduction of SVD results in an almost 70\% improvement in FVD. Additionally, we also note that a slight decrease in FID, which we hypothesize is attributed to the introduction of the cross-view module, disrupting the SVD's ability to learn spatial features. To fully unleash the potential of SVD in multi-view video generation, we propose UniMVM, which unifies constraints on intra- and cross-view, achieving remarkable FID and FVD scores of 11.2 and 55.7, respectively. These represent relative improvements of $\sim$30\% and $\sim$80\% compared to DriveDreamer. As depicted in Fig. \ref{fig_mv_1}, the upper row in each compared pair is generated by \textit{DriveDreamer-2} without UniMVM, while the lower row is generated by \textit{DriveDreamer-2} with UniMVM. In the absence of UniMVM, \textit{DriveDreamer-2} generates inconsistent results between views, including foreground vehicles and background structures. The introduction of UniMVM leads to significant improvements in generating multi-view videos, both in foreground and background aspects. The qualitative results demonstrate the impressive capability of our UniMVM in achieving multi-view consistency. 

\begin{table}[t]
    \begin{center}
    \caption{The ablation study on the backbone and UniMVM. Cross-view module denotes the cross-view attention used in previous methods\cite{wang2023drivedreamer,wang2023driving,wen2023panacea,drivediff}. }\label{tab_ablation}
    \setlength{\tabcolsep}{1.7mm}
    \begin{tabular}{cccccc}
        \hline  
        Method&Backbone&Cross-view module&UniMVM&FID$\downarrow$ &FVD$\downarrow$ \\
        \hline 
        DriveDreamer \cite{wang2023drivedreamer}&SD1.4&\checkmark&-&14.9&340.8\\
        \hline
        \multirow{2}*{DriveDreamer-2}&\multirow{2}*{SVD}&\checkmark&-&17.2&94.6\\
        ~&~&-&\checkmark&\textbf{11.2}&\textbf{55.7}\\
        \hline
    \end{tabular}
    \vspace{-1em}
    \end{center}
\end{table}

Moreover, we explore the influence of various conditions on driving video generation, as shown in Tab.~\ref{tab_ablation_cond} and Fig.~\ref{fig_mv}. The first row in Fig.~\ref{fig_mv} illustrates the ground truth (GT) of the first frame, representing the style of the GT video. Meanwhile, the second row displays the GT of the second frame, representing the GT of the generated multi-view frame. \textit{DriveDreamer-2} with the initial frame can generate results that are highly similar to the GT video, achieving optimal results in terms of FID and FVD, with scores of 11.2 and 55.7, respectively. The video generated by \textit{DriveDreamer-2} with the front-view video retains some aspects of the GT scene while also introducing some diversity, resulting in 17.2 FID and 94.6 FVD. \textit{DriveDreamer-2} can also generate extremely competitive results even without any image conditioning, achieving FID and FVD scores of 25.0 and 105.1, respectively. Notably, \textit{DriveDreamer-2} exhibits
the highest diversity in this setting, where the generated appearance of cars and the street backgrounds differ significantly from the ground truth.

\begin{table}[t]
    \begin{center}
    \caption{The ablation study on different conditions.}\label{tab_ablation_cond}
    \setlength{\tabcolsep}{1.7mm}
    \begin{tabular}{cccc}
        \hline  
        Method&Conditions&FID$\downarrow$ &FVD$\downarrow$ \\
        \hline
        \multirow{2}*{DriveDreamer \cite{wang2023drivedreamer}}&-&26.8&353.2\\
        ~&1st-frame multi-view image&14.9&340.8\\
        \hline
        \multirow{3}*{DriveDreamer-2}&-&25.0&105.1\\
        ~&1-view video&18.4&74.9\\
        ~&1-st frame multi-view image&\textbf{11.2}&\textbf{55.7}\\
        \hline
    \end{tabular}
    \vspace{-3em}
    \end{center}
\end{table}

\section{Discussion and Conclusion}
This paper introduces \textit{DriveDreamer-2}, an innovative extension of the DriveDreamer framework that pioneers the generation of user-customized driving videos. Leveraging a Large Language Model, \textit{DriveDreamer-2} first transfers user queries into foreground agent trajectories. Then the background traffic conditions can be generated using the proposed HDMap generator, with agent trajectories as conditions. The generated structured conditions can be utilized for video generation, and we propose UniMVM to enhance temporal and spatial coherence.
We conduct extensive experiments to verify that \textit{DriveDreamer-2} can generate uncommon driving videos, such as abrupt vehicle maneuvers. Importantly, experimental results showcase the utility of the generated videos in enhancing the training of driving perception methods. Furthermore, \textit{DriveDreamer-2} demonstrates superior video generation quality compared to state-of-the-art methods, achieving FID and FVD scores of 11.2 and 55.7, respectively. These scores represent remarkable relative improvements of approximately 30\% and 50\%, affirming the efficacy and advancement of \textit{DriveDreamer-2} in multi-view driving video generation.
\vspace{-0.5em}
{\small
\bibliographystyle{splncs04}
\bibliography{ref}
}

\newpage

In the supplementary material, we begin by detailing the implementation of customized traffic simulation and the Unified Mulit-view Video Model (UniMVM). This is followed by an overview of the training specifics for downstream tasks and evaluation details. Additionally, a collection of visualization results is provided for further analysis.
\section{Implement Details}
\noindent{\textbf{Function Library}} Below is an example of a \textit{cut in} trajectory generation function. Firstly, we initialize a starting point coordinate, and then during the generation of the \textit{cut in} trajectory, we check the distance between the agent and other agents to control direction and avoid collisions. To enhance the diversity of generated trajectories, we introduce random disturbances to the agent’s speed and orientation, aiming to approximate real-world scenarios as closely as possible. Moreover, this function library can be further expanded to meet users' demands for generating scenes. More details about the function library can be found in the code that we will release later.

\begin{python}
def cut_in(obj_trajs=None,obj_vels=None,
            safe_dis=10, is_ego=False):
    '''
    Generate a trajectory for a agent cutting in.
    Parameters:

    Returns:
    obj_trajs: the cutting in trajectory of the agent.
    obj_vels: the velocities of the agent.
    '''
    agent_trajs = np.zeros((1,NUM_POINT,3))
    agent_vels = np.zeros((1,NUM_POINT,1))
    #initialize the start point
    if not is_ego:
        xy0 = np.random.rand(2)
        xy0[0] = utils.get_random_value(xy0[0],[0,10])
        multi_factor = 1 if random.random()>0.5 else -1
        xy0[1] = utils.get_random_value(xy0[1],Y_RANGE)\
                *multi_factor
        y_margin = xy0[1]*multi_factor 
        agent_trajs[0,0,:2] = xy0
        target_y = 0
    else:
        y_margin = abs(obj_trajs[0,0,1])
        target_y=obj_trajs[0,0,1]
    vel = utils.get_random_value(np.random.rand(),V_RANGE)
    agent_vels[0,0] = vel
    yaw = utils.get_random_value(np.random.rand(),FORWARD_RANGE)
    vel_x = vel*np.cos(yaw)
    vel_y = vel*np.sin(yaw)
    flag = True
    # generate the cut in trajectory
    for t in range(1,NUM_POINT):
        agent_trajs[0,t,0] = agent_trajs[0,t-1,0]\
                             +vel_x*T_INTER 
        agent_trajs[0,t,1] = agent_trajs[0,t-1,1]\
                             +vel_y*T_INTER 
        #check whether the agent reaches the front of the 
        #target vehicle
        if (agent_trajs[0,t-1,1]-y_margin)*\
           (agent_trajs[0,t,1]-y_margin)<=0:
            flag = False
        # approach the target vehicle
        if abs(agent_trajs[0,t,1]-target_y)\
           >y_margin/2 and flag:
            yaw_range = [-0.1,0] if multi_factor==1 \
                        else [0,0.1]
            yaw_max = 0 if multi_factor==1 else 20*np.pi/180
            yaw_min = -20*np.pi/180 if multi_factor==1 else 0
        # gradually return to the forward direction.
        elif abs(agent_trajs[0,t,1]-target_y)\
             >0.5 and flag:
            yaw_range = [0,0.1] if multi_factor==1\ 
                        else [-0.1,0]
            yaw_max = -10*np.pi/180 if multi_factor==1\ 
                     else 20*np.pi/180
            yaw_min = -20*np.pi/180 if multi_factor==1\ 
                      else 10*np.pi/180
        # move forward
        else:
            yaw_range = [-0.3*np.pi/180,0] if yaw>=0 \
                        else [0,0.3*np.pi/180]
            yaw_max = 1*np.pi/180 
            yaw_min = -1*np.pi/180 
        # finetune the direction to avoid collision
        if obj_trajs is not None:
            yaw_range = utils.update_yaw(yaw,safe_dis,agent_trajs,obj_trajs)
        yaw += utils.get_random_value(np.random.rand(),yaw_range)
        yaw = np.clip(yaw,yaw_min,yaw_max)
        vel += utils.get_random_value(np.random.rand(),[-2,2])
        # guarentee the generated velocity surpasses the
        # other vehicle to relize cutting in
        if obj_vels is not None:
            v_range = [obj_vels[0,t,0]+0.5,V_RANGE[1]+0.5]
        else:
            v_range = V_RANGE
        vel = np.clip(vel,v_range[0],v_range[1])
        agent_vels[0,t] = vel
        vel_x = vel*np.cos(yaw)
        vel_y = vel*np.sin(yaw)
    return agent_trajs,agent_vels
    
\end{python}

\begin{figure}[h]
\vspace{-1em}
    \centering
    \includegraphics[width=\textwidth]{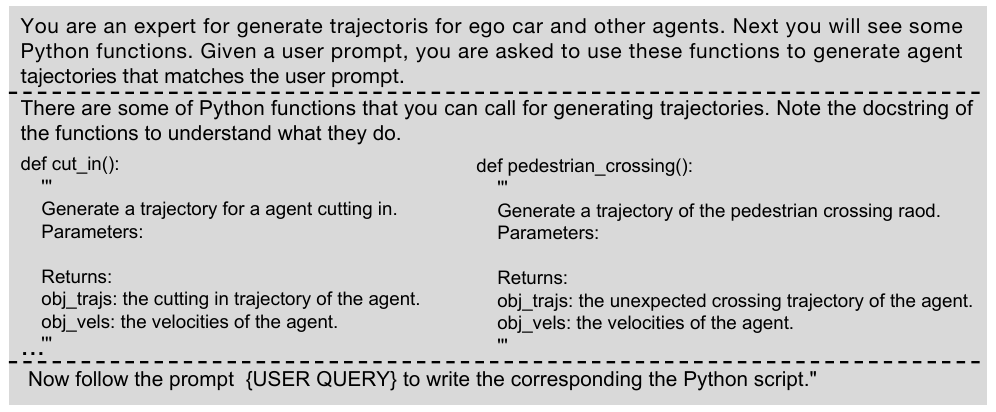}
    \caption{Our prompt template designed for finetuning LLM. It contains information about functions, and instruction. The user query is inserted into the placeholder \{USER QUERY\}.}
    \label{prompt_template}
    \vspace{-1em}
\end{figure}

\noindent{\textbf{Prompt Template}} Fig.~\ref{prompt_template} depicts the general prompt template designed for finetuning LLM. It includes encapsulated trajectory generation functions, and instruction. During usage, the user prompt integrates into the instruction, directly guiding the finetuned LLM to generate the corresponding Python script. 

\noindent{\textbf{BEV HDMap Post-Process}} Since the generated HDMap is in the BEV perspective, it cannot be directly applied to UniMVM for video generation. Hence, we need post-processing to project the BEV HDMap onto the image coordinate system. Firstly, binarization and skeleton extraction algorithms are employed to extract different types of lane markings (as depicted in Fig.~4 (in the main text), {\color{red}red} for lane boundaries, {\color{blue}blue} for lane dividers, and {\color{green}green} for pedestrian crossings). Then, the extracted lane markings are transformed into pixel coordinates to serve as the condition for UniMVM.

\begin{figure}[t]
    \centering
    \includegraphics[width=\textwidth]{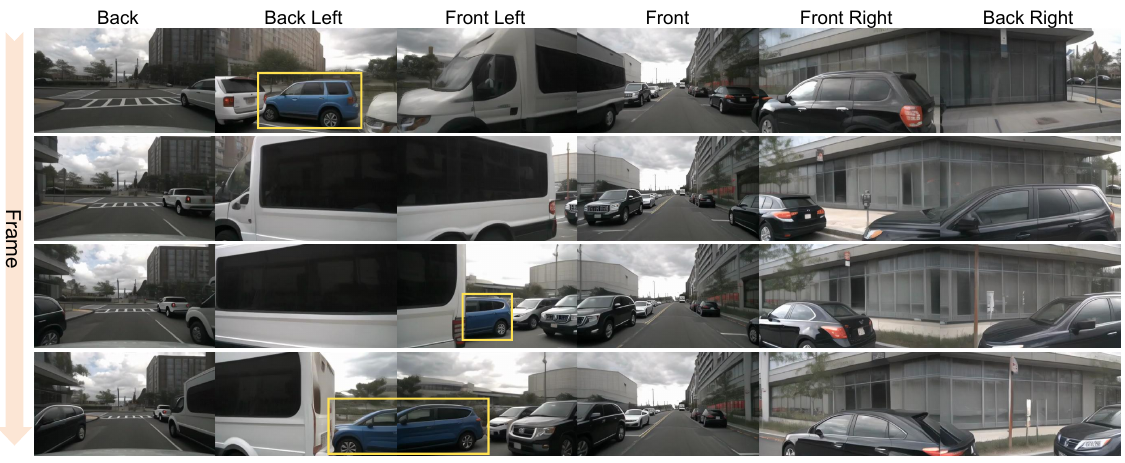}
    \caption{Visualization of the generated multi-view video. Regions highlighted by {\color{yellow}yellow} rectangles indicate that \textit{DriveDreamer-2} exhibits strong temporal consistency.}
    \label{block}
\end{figure}

\begin{figure}[h]
    \centering
    \includegraphics[width=\textwidth]{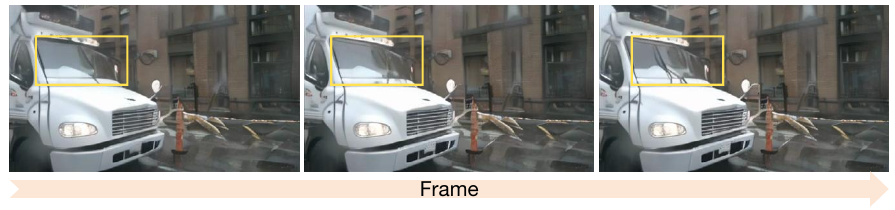}
    \caption{A generated video of a rainy day scene. The movement of the windshield wipers (highlighted by {\color{yellow}yellow} rectangles) demonstrates that \textit{as DriveDreamer-2} possesses a high level of scene understanding.}
    \label{wiper}
    \vspace{-2em}
\end{figure}

\noindent{\textbf{Training UniMVM}} We assume the data distribution is $p_{data}(y_{\cal I})$, where $y_{\cal I}$ denotes the image latents, and let $p(y;\sigma)$ be the distribution obtained by adding i.i.d. $\sigma^2$-variance Gaussian noise to the data. Note that or sufficiently large $\sigma_{max}$, $p(y;\sigma_{max^2}) \approx {\cal N}(0,\sigma_{max^2})$. Diffusion models (DM) \cite{df,df1,df2,df3} use this fact and, starting from high variance Gaussian noise $y_M\sim{\cal N}(0,\sigma_{max^2})$, sequentially denoise towards $\sigma_0=0$. In practice, this iterative refinement process
can be implemented through the numerical simulation of the Probability Flow ordinary differential equation (ODE) \cite{song2020score}
\begin{equation}
dx=-\dot{\sigma}(t)\sigma(t)\nabla_y \log p(y; \sigma(t)) dt,
\end{equation}
where $\nabla_y \log p(y;\sigma)$ is the score function \cite{hyvarinen2005estimation}. DM training reduces to learning a model $s_\theta(x; \sigma)$ for the score function $\nabla_x \log p(y;\sigma)$. 
The model can, for example, be parameterized as $\nabla_x \log p(y;\sigma)\approx s_\theta(y; \sigma) = (D_\theta(y;\sigma)-y)/\sigma^2$ \cite{edm}, where $D_\theta$ is a learnable denoiser that tries to predict the clean $y_{\cal I}$. The denoiser $D_\theta$ is trained via denoising score matching (DSM)
\begin{equation}
    \mathbb{E}_{(y_{\cal I},c)\sim p_{data}(y_{\cal I},c),(\sigma,n)\sim p(\sigma,n)}\left[\lambda_\sigma\Vert D_\theta(y_{\cal I}+n;\sigma,c)-y_{\cal I}\Vert_2^2\right],
\end{equation}
where $p(\sigma,n)=p(\sigma){\cal N}(n;0,\sigma^2)$, $p(\sigma)$ can be a probability distribution or density over noise levels $\sigma$. $\lambda_\sigma=(1+\sigma^2)\sigma^{-2}$ is a weighting function, and $c$ is the conditional signal, which including the HDMap features $y_{\cal H}$, box conditions $y_{\cal B}$ and text prompt embeddings $c_p$. In this work, we follow the EDM preconditioning framework \cite{edm}, parameterizing the learnable denoiser $D_\theta$ as
\begin{equation}
    D_\theta(y; \sigma) = c_{skip}(\sigma)y + c_{out}(\sigma)F_\theta(c_{in}(\sigma)y; c_{noise}(\sigma)),
\end{equation}
where $F_\theta$ is the network to be trained, and $c_{skip}(\sigma)=1/(\sigma^2+1), c_{out}(\sigma)=-\sigma/\sqrt{\sigma^2+1}$, $c_{in}(\sigma)=1/\sqrt{\sigma^2+1}$, $c_{noise}(\sigma)=0.25\log\sigma$.

\noindent{\textbf{Training Downstream Tasks}} StreamPETR is retrained at a resolution of $512\times256$ instead of the $704\times256$ in original baseline \cite{wang2023exploring}. The augmented training data is generated by utilizing the structured information from the training set.

\noindent{\textbf{Evaluation Details}} The FID and FVD calculations are performed on 150 validation videos from the nuScenes dataset \cite{nusc}. A setting with a frame rate of 4Hz and 8 frames per clip is employed to generate data for evaluating FID and FVD. And we use the official UCF FVD evaluation code\footnote{\url{https://github.com/SongweiGe/TATS/}}.

\begin{figure}[t]
    \vspace{-2em}
    \centering
    \includegraphics[width=\textwidth]{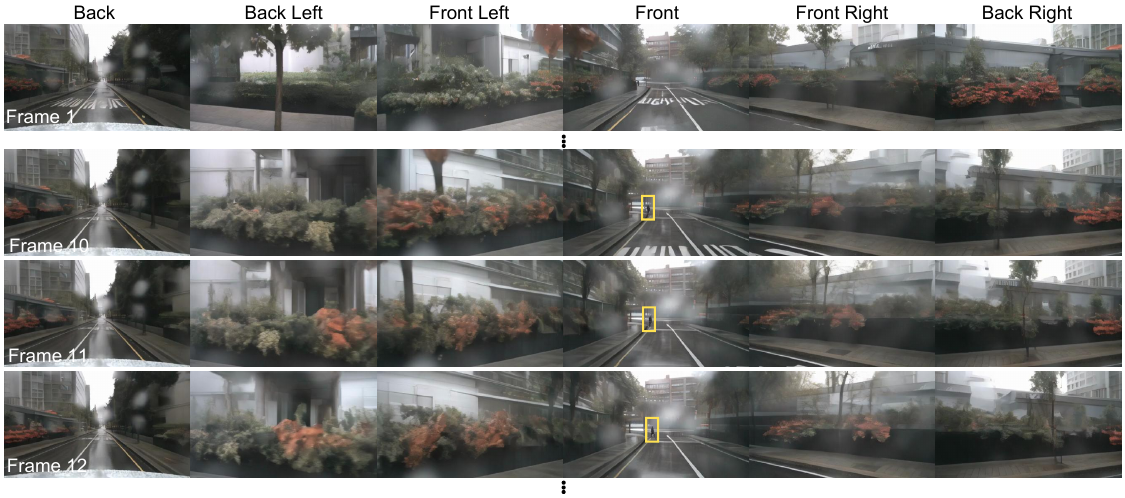}
    \caption{A long video generated by the prompt \textit{a person crosses the road on a rainy day}. Regions highlighted by {\color{yellow}yellow} rectangles indicate the movement of the pedestrian.}
    \label{crossing_road}
    \vspace{-2em}
\end{figure}

\section{Visualization}

\begin{figure}[ht]
    \vspace{-2em}
    \centering
    \includegraphics[width=\textwidth]{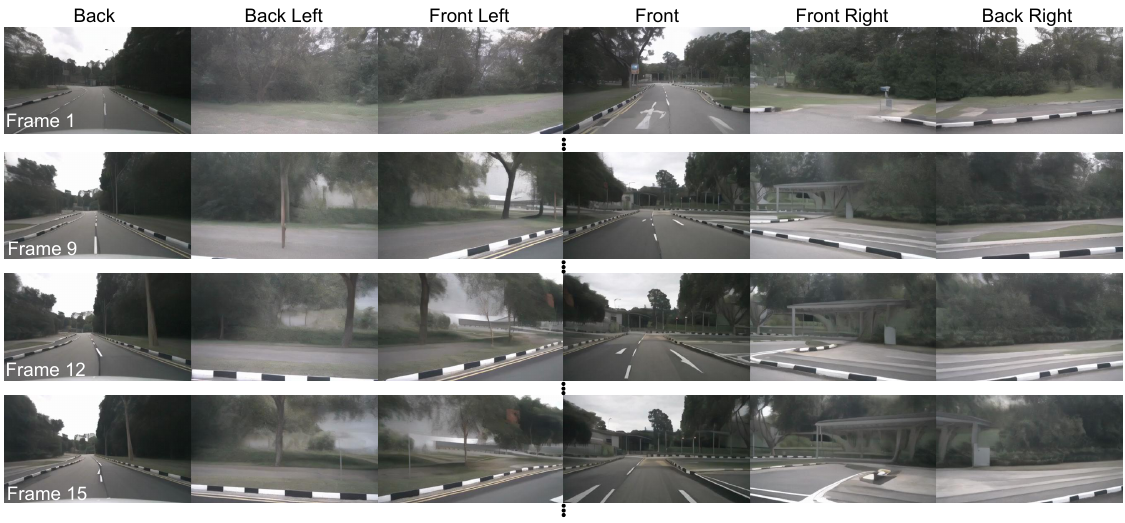}
    \caption{A long video generated by the prompt \textit{the ego car changes lane during the daytime}.}
    \label{change_lane}
    \vspace{-2em}
\end{figure}

As shown in Fig.~\ref{block}, \textit{DriveDreamer-2} demonstrates strong consistency across views and frames, allowing even the completely occluded car to reappear in subsequent frames (as highlighted by {\color{yellow}yellow} rectangles). The more details can be seen in videos/occluded\_car.mp4. 
Fig.~\ref{wiper} depicts a rainy scene video generated by \textit{DriveDreamer-2} (see videos/windshield\_wiper.mp4 for more details). As illustrated in Fig.~\ref{wiper}, the windshield wipers of the truck are continuously clearing the windshield. This showcases the powerful scene understanding capabilities of our \textit{DriveDreamer-2}. It can not only manipulate macroscopic weather conditions but also adjust the behaviors of generated agents within the scene according to the weather.

\begin{figure}[ht]
    \vspace{-1em}
    \centering
    \includegraphics[width=\textwidth]{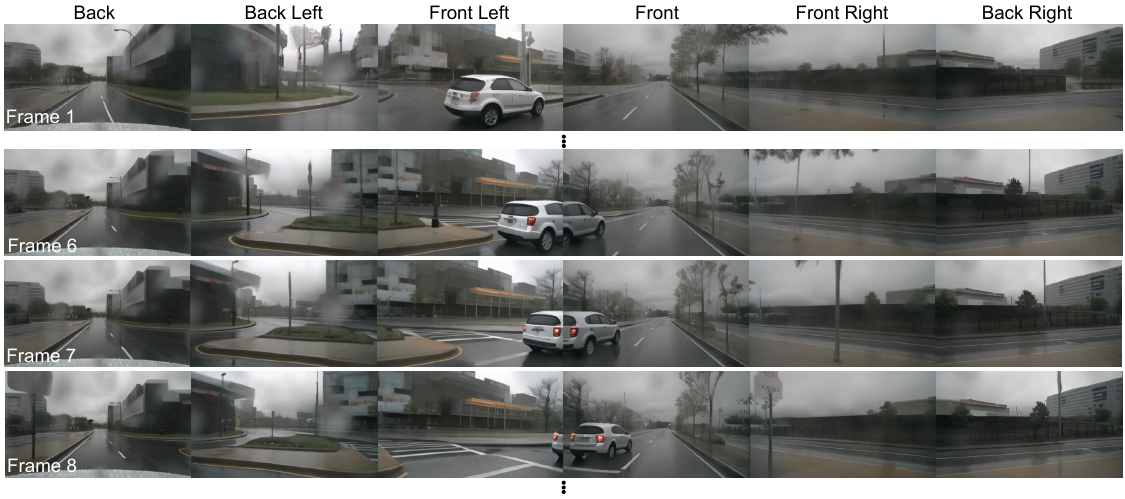}
    \caption{A long video generated by the prompt \textit{on a rainy day, a car cuts in}.}
    \label{cut_in}
    \vspace{-2em}
\end{figure}

Regarding the generation of long videos, we can utilize \textit{DriveDreamer-2} without any image conditioning to first generate a clip. Subsequently, the last frame of this clip is employed as the initial image condition to generate subsequent clips. Below, we showcase the results of directly using text prompts to generate a 29-frame multi-view video. Fig.~\ref{crossing_road} is a long video generated by the prompt \textit{a person crosses the road on a rainy day}. (see videos/cross\_road.mp4 for more details). It depicts a pedestrian crosses the road in front of the ego vehicle on a rainy day. Fig.~\ref{change_lane} is generated by a prompt \textit{the ego car changes lane during the daytime}. (see videos/change\_lane.mp4 for more details). It showcases the ego car changes lanes to the right side during the daytime. As shown in Fig.~\ref{cut_in}, a vehicle cuts in from the left to the front of the ego car (prompted by \textit{on a rainy day, a car cuts in} and see more details in videos/cut\_in.mp4).

\end{document}